\documentclass[letterpaper]{article} % DO NOT CHANGE THIS
\usepackage[submission]{aaai23}  % DO NOT CHANGE THIS
\usepackage{times}  % DO NOT CHANGE THIS
\usepackage{helvet}  % DO NOT CHANGE THIS
\usepackage{courier}  % DO NOT CHANGE THIS
\usepackage[hyphens]{url}  % DO NOT CHANGE THIS
\usepackage{graphicx} % DO NOT CHANGE THIS
\usepackage{natbib}  % DO NOT CHANGE THIS AND DO NOT ADD ANY OPTIONS TO IT
\usepackage{caption} % DO NOT CHANGE THIS AND DO NOT ADD ANY OPTIONS TO IT
\usepackage{algorithm}
\usepackage{algorithmic}
\usepackage{amsmath, amsfonts, amssymb}
\usepackage{mathrsfs}
\usepackage{newfloat}
\usepackage{listings}
\usepackage{booktabs}
\usepackage{multirow}
\DeclareCaptionStyle{ruled}{labelfont=normalfont, labelsep=colon, strut=off} % DO NOT CHANGE THIS
\floatstyle{ruled}
\newfloat{listing}{tb}{lst}{}
\floatname{listing}{Listing}
\title{Look in Different Views: Multi-Scheme Regression Guided Cell Instance Segmentation}
\author{
    Written by AAAI Press Staff\textsuperscript{\rm 1}\thanks{With help from the AAAI Publications Committee. }\\
    AAAI Style Contributions by Pater Patel Schneider, 
    Sunil Issar, \\
    J. Scott Penberthy, 
    George Ferguson, 
    Hans Guesgen, 
    Francisco Cruz\equalcontrib, 
    Marc Pujol-Gonzalez\equalcontrib
}
\affiliations{
    \textsuperscript{\rm 1}Association for the Advancement of Artificial Intelligence\\
    1900 Embarcadero Road, Suite 101\\
    Palo Alto, California 94303-3310 USA\\
    publications23@aaai. org
}

\begin{document}

\maketitle

\begin{abstract}
Cell instance segmentation is a new and challenging task aiming at joint detection and segmentation of every cell in an image. Recently, many instance segmentation methods have applied in this task. Despite their great success, there still exists two main weaknesses caused by uncertainty of localizing cell center points. First, densely packed cells can easily be recognized into one cell. Second, elongated cell can easily be recognized into two cells. To overcome these two weaknesses, we propose a novel cell instance segmentation network based on multi-scheme regression guidance. With multi-scheme regression guidance, the network has the ability to look each cell in different views. Specifically, we first propose a gaussian guidance attention mechanism to use gaussian labels for guiding the network's attention. We then propose a point-regression module for assisting the regression of cell center. Finally, we utilize the output of the above two modules to further guide the instance segmentation. With multi-scheme regression guidance, we can take full advantage of the characteristics of different regions, especially the central region of the cell. We conduct extensive experiments on benchmark datasets, DSB2018, CA2.5 and SCIS. The encouraging results show that our network achieves SOTA (state-of-the-art) performance. On the DSB2018 and CA2.5, our network surpasses previous methods by 1.2\% (AP50). Particularly on SCIS dataset, our network performs stronger by large margin (3.0\% higher AP50). Visualization and analysis further prove that our proposed method is interpretable.
\end{abstract}
\section{Introduction}
With the rapid development of instance segmentation task, cell instance segmentation gradually becomes an promising application in modern medical treatment. This application aims at joint detection and segmentation of every cell in an image. There are two widely used imaging methods for living cells in medical application: fluorescence techniques and non-invasive techniques. Image boundaries using fluorescence techniques are clearer, while non-invasive techniques allows imaging cells without damaging the cell structure.
\begin{figure}[t]
\centering
\includegraphics[scale=0.69]{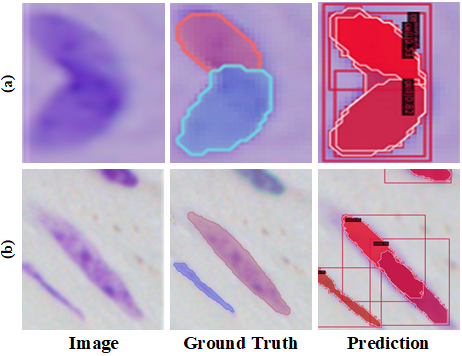}
\caption{Intuitive cases to explain the two main weaknesses in cell instance segmentation. (a) A typical case that densely packed cells are recognized into one cell.(b) A typical case that elongated cells are recognized into two cells. }
\label{fig:failurecases}
\end{figure}
\par Recently, many instance segmentation methods have made great success in this field~\cite{DBLP:conf/isbi/BouyssouxFO22, DBLP:conf/isbi/YiWHQHM19}. The existing notable instance segmentation methods can be divided into two categories: single-stage and multi-stage methods. The multi-stage methods decompose the instance segmentation task into two parts: object detection and mask segmentation~\cite{He_2017_ICCV, Liu_2018_CVPR}. The single-stage methods integrate detection and segmentation into a whole architecture, where some methods are followed single-stage target detection networks~\cite{wang2020solo, yolact-iccv2019} and others are inspired by anchor-free detection models~\cite{Xie_2020_CVPR, DBLP:journals/corr/abs-1909-07829}. Even though huge success has been achieved, there still exists two main weaknesses shown in Fig. \ref{fig:failurecases}. Intuitively, densely packed cells be recognized into one cell (Fig.\ref{fig:failurecases} (a)) and elongated cells be recognized into two cells (Fig.\ref{fig:failurecases} (b)).
\par
\begin{figure*}[ht]

\centering
\includegraphics[scale=0.37]{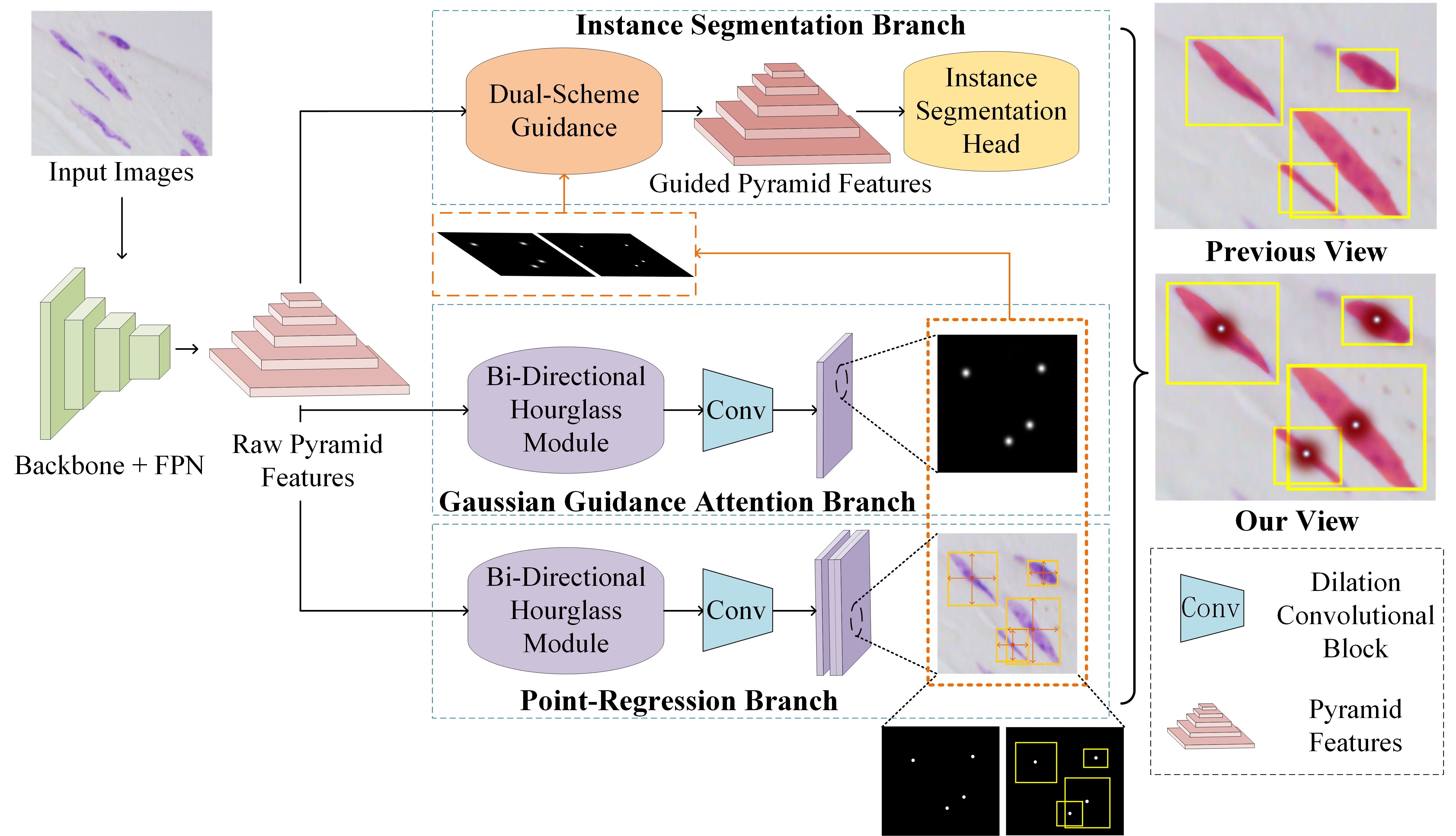}
\caption{The framework of our proposed MSRNet. MSRNet is composed of three main branches: instance segmentation branch, gaussian guidance attention branch and point-regression branch. The prediction masks of gaussian guidance attention branch and point-regression branch will be sent to the dual-scheme guidance module for the final instance predictions. It is worth noting that the instance segmentation head is based on the well-known SCNet\cite{scnet_2021_aaai}}
\label{fig:maingraph}
\end{figure*}

 In this paper, we propose a multi-scheme regression guided cell instance
segmentation network, MSRNet, to address the weaknesses above. By applying multi-scheme regression, MSRNet makes it easier to regress the correct cell center location. Since MSRNet pays more attention to the center of the cell, the discriminating ability between different individual cells has been improved. As shown in Fig. \ref{fig:maingraph}, we propose a gaussian guidance attention branch (GGAB) and point-regression branch (PRB) alongside with instance segmentation branch. Specifically, GGAB predicts the approximate location of the cell center on the image. In this branch, a bi-directional hourglass module is used to make the output features have both the detailed lower level features and necessary higher level features. gaussian labels are used to supervise the learning. With this design, the weakness shown in Fig.\ref{fig:failurecases} (b) can be maximally eased, because the instance segmentation branch focus more on the regions around cell center so that one cell is not easily be recognized into two. PRB locates the cell centers in a fine-grained manner. In this branch, after a bi-directional hourglass module, the point-wise masks and the height and width information of each cell are used to supervise the learning. With this design, the weakness shown in Fig.\ref{fig:failurecases} (a) can be maximally eased, because gaussian mask can't easily recognize densely packed cells, while more fine-grained point-regression branch can correctly localize these cells' center points. Besides, we present a dual-scheme guidance module before instance segmentation head. This module could guide the instance segmentation branch with two types of masks from the output of GGAB and PRB. Compared to previous method, our proposed MSRNet regress the cell position from multiple schemes, in order to correctly locate each cell center. Basically, our proposed MSRNet can look cells in different views. Extensive experiments demonstrate that MSRNet achieves SOTA performance on DSB2018, CA2.5~\cite{ca2.5_10.1007/978-3-030-87237-3_43} and SCIS~\cite{edlund2021livecell} datasets. Noteworthy, MSRNet performs stronger by large margin (3.0\% higher AP50) on SCIS dataset. Visualization and analysis further demonstrate the interpretability of our proposed MSRNet. The contributions of this paper are summarized as follows:
\begin{itemize}
    \item We propose a gaussian guidance attention branch to make the instance segmentation branch focus on the regions around cell center.
    \item We propose a point-regression branch in order to enhance the regression ability of MSRNet to locate the cell center in a fine-grained manner, which can further distinguish densely packed cells.
    \item We present a dual-scheme guidance module to guide the instance segmentation branch receive from two types of masks from the output of the other two branches.
    \item Encouraging experimental results show that MSRNet achieves SOTA performance on the DSB2018, CA2.5~\cite{ca2.5_10.1007/978-3-030-87237-3_43} and SCIS datasets~\cite{edlund2021livecell}, which prove the effectiveness of MSRNet.
\end{itemize}

\section{Related Work}
\subsection{Medical Image Instance Segmentation}
There are currently two main categories for instance segmentation: single-stage and multi-stage methods. The process of multi-stage methods is to first find out the bounding boxes by means of object detection, then pixel-wise segmentation is been performed within the proposed regions. Mask R-CNN~\cite{He_2017_ICCV} is the most famous method under this approach, which merges a mask branch parallel to the bounding boxes branch for Faster R-CNN~\cite{DBLP:journals/corr/RenHG015}. A large number of Mask R-CNN based instance segmentation works have been proposed in recent years. For example, Cascade Mask R-CNN~\cite{Cai_2019} uses cascade regressors and detectors to enable instance segmentation head to better fit the data; Hybrid Task Cascade (HTC)~\cite{chen2019hybrid} crosses the bounding box regression branch and mask prediction branch, and use information from semantic labels to guide instance segmentation. Sample Consistency Network (SCNet)~\cite{scnet_2021_aaai} improves the network structure of Cascade Mask R-CNN to ensure that the IoU distribution in training time and inference time is close. In recent years, many excellent works relate to single-stage instance segmentation have been published, some of these methods are faster and more accurate than Mask R-CNN. For example, SOLO~\cite{wang2020solo} transforms the instance segmentation task into a classification task. The main method of SOLO is to divide the image into S*S regions and use two branches to predict category and mask of each instance. SOLOv2~\cite{wang2020solov2} further enhances the performance of SOLO.
\par For medical image instance segmentation tasks, especially cell instance segmentation tasks, the difficulties encountered in segmentation are mainly the dense distribution of instances, unclear instance boundaries and the variability in the appearance of different instances\cite{DBLP:conf/eccv/YiWJHM18, DBLP:conf/isbi/YiWHM18}. These cause difficulties for the regression of cell locations. For this task, many specialized instance segmentation networks are proposed. For example, DCAN~\cite{Li_Liu_Lin_Xie_Ding_Huang_Tang_2020} is an efficient net which uses two different branches to output the results of object detection and contours segmentation. STARDIST~\cite{10.1007/978-3-030-00934-2_30} uses the prior that cells are all convex polygon shapes for segmentation, which requires that the cells in the dataset be convex. CosineEmbedding~\cite{10.1007/978-3-030-00934-2_1} recognizes cells by clustering, while it tends to produce a large number of false positives due to the different clustering results of each cell.
\subsection{Point-guided Attention Mechanism}
The attention mechanism is a module inspired by human visual mechanism that helps model to focus on the more critical parts of the input image. Attention mechanism was first proposed in a language translation task~\cite{DBLP:conf/emnlp/ChoMGBBSB14}, now it has been used for all aspects of tasks including classification~\cite{DBLP:conf/naacl/YangYDHSH16}, object detection~\cite{DBLP:journals/corr/abs-1804-05338}, and image segmentation~\cite{DBLP:journals/corr/abs-1804-03999} and can be divided into two categories: soft attention and hard attention~\cite{DBLP:conf/icml/XuBKCCSZB15}. Guided Attention Inference Network (GAIN)~\cite{8733010} explains what the learner focus on, and can feeds back with direct guidance towards speciﬁc tasks. Mask-guided Contrastive Attention Model~\cite{DBLP:conf/cvpr/Song0O018} helps the feature extractor focus on body region instead of background region, so that the network will be more adaptable to the background. In this paper, we introduce a gaussian cell center annotation to supervise the gaussian guidance attention branch. We take inspiration of some networks based on gaussian guidance. MCNN~\cite{DBLP:conf/cvpr/ZhangZCGM16} is proposed to regress gaussian maps for different head sizes with multi-column convolutions. CSRNet~\cite{DBLP:conf/cvpr/LiZC18} uses a series of dilated convolutional blocks to have a large field of perception and be able to extract deep features. Ma \emph{et al. } present a bayesian loss~\cite{DBLP:journals/corr/abs-1908-03684} which directly use the point annotations as supervision, use expectations instead of pseudo labels to calculate loss. We take advantage of such methods and propose a gaussian guidance attention branch for instance segmentation.
\section{Method}
In this paper, we propose MSRNet for cell instance segmentation. With multi-scheme regression mechanism, the weaknesses encountered (Fig.\ref{fig:failurecases}) in cell instance segmentation will be alleviated. The overview of our proposed network is illustrated in Fig. \ref{fig:maingraph}. Generally, there are three branches, which are instance segmentation branch, gaussian guidance attention branch and point-regression branch.
\subsection{Gaussian Guidance Attention Branch}
Single cell recognized as multiple cells is a serious weakness in cell instance segmentation. When cells are elongated, they are easily to be incorrectly recognized into two separated smaller cells. Aiming at this weakness, we propose a gaussian guidance attention branch. During training phase, this attention branch will learn to generate gaussian prediction masks for cell images. With optimization, our backbone will pay more attention on the center region of cells. On this branch, we first use the instance segmentation label to construct a gaussian mask for each input image. In this paper, we use the object center positions to construct our gaussian masks. Specifically, around the center position, we generate each cell's mask processed by a gaussian kernel. We treat the upper limit of the gaussian radius as a hyperparameter in this paper. When the length of the bounding box's short edge is shorter than this gaussian radius, this short edge's length will be used as the new gaussian radius to prevent the gaussian circle from overflowing the bounding box range. 
\par In general, low-level features contain more detailed information, which is urgently needed for instance segmentation of densely packed cells. To take full advantage of the multi-level features, we present a bi-directional hourglass module. In this module, a bottom-up fusion is first applied to the pyramid features to make each layer of features obtain detailed features. Then the highest level feature map is processed by multiple up-samplers. After each up-sampling, a channel-wise concatenation is fused with the features at same stage and previous stage (requires an additional down-sampling). This module enables the output features to preserve as much detailed information as possible and necessary high level feature map's information. Some dilation convolutional blocks are then applied to the output feature. Finally, we adopt a MSE (Mean-Square Error) loss to supervise the learning using the gaussian mask.
\begin{equation}
\label{module_first}
\begin{aligned}
&F_{i+1}^{mid}=f_{i}^{mid}(Concat(F_{i+1}^{in}, Pool(F_{i}^{mid})))\\
&F_{i}^{out}=f_{i}^{out}(F_{i}^{mid}+\beta_{i}Pool(F_{i-1}^{mid})+\gamma_{i}Up(F_{i+1}^{out}))\\
\end{aligned}
\end{equation}
where $f$ means convolutional blocks while $\beta_{i}$ and $\gamma_{i}$ are learnable parameters that control the proportion learn from each input feature map. $Up$ indicates upsamplers and $i = 1, 2, 3, \cdots$.
\par To solve the weakness (b) showing in Fig. \ref{fig:failurecases}, we propose GGAB to improve the network's perception of the cell's central region. This module can help the feature map to generate sufficient activation in central region, thus prevent the cells from being truncated.
\begin{figure}[tbp]
\centering
\includegraphics[scale=0.22]{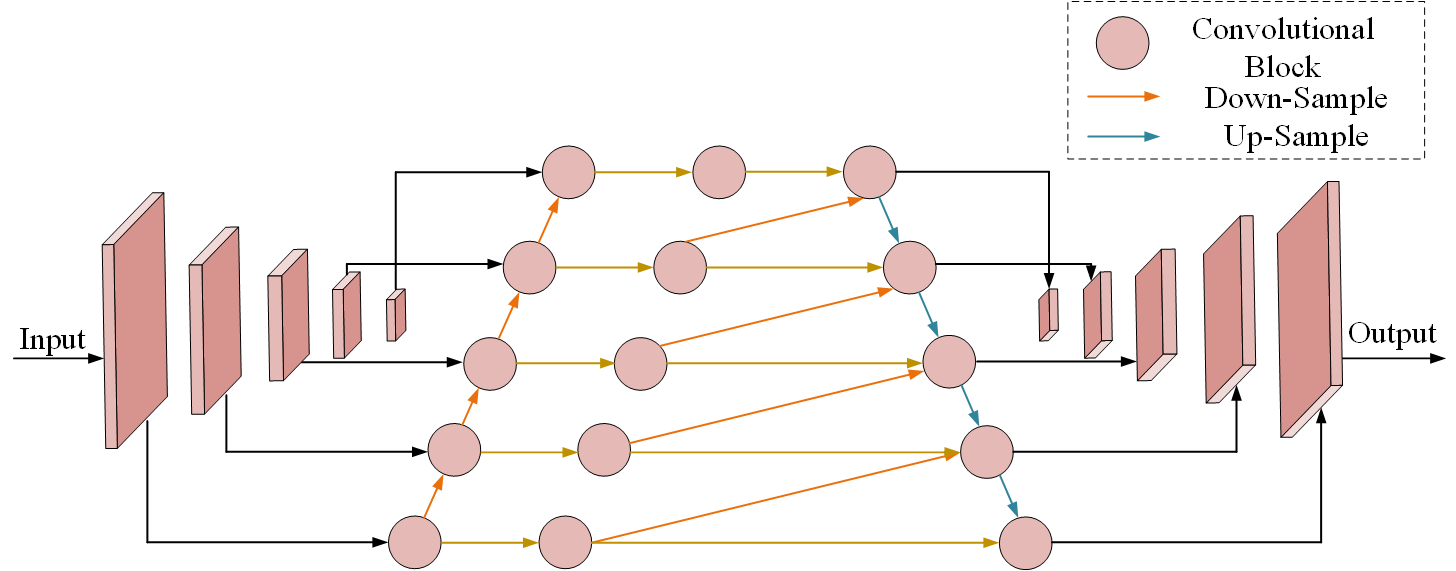}
\caption{Bi-Directional Hourglass Module}
\label{fig:multiscale}
\end{figure}
\subsection{Point-Regression Branch}Paralleled to GGAB, we propose a point-regression branch. Similar to GGAB, raw pyramid features is processed by a bi-directional hourglass module. Then, passed through some convolutional blocks, two types of masks are generated, which are center point masks and height-width predictions. These should be supervised by two types of ground truths constructed by instance segmentation labels. The center point mask has a non-zero value only at the cell-centered pixels. The height-width label is a 2-channel tensor that contains the height and width data of each boundary boxes of instance segmentation labels. Only the center point masks are used to guide the instance segmentation branch in the dual-scheme guidance module, while the height-width predictions are only used to assist the PRB to regress the correct cell center's position.
\par By applying this module, we aim to solve the weakness (a) showing in Fig. \ref{fig:failurecases}. When two cells are densely packed, gaussian masks may have difficulty distinguishing between these two cells, because predicted gaussian masks may also be densely packed. Under this situation, instance segmentation branch may recognize these two cells as one with confusing guidance. On solving this weakness, we propose point regression branch to regress the location of cells in a fine-grained manner.
\par Although PRB is effective, using PRB alone without GGAB is far less effective than using two branches at the same time. Because PRB cannot generate enough activation in the central region, using PRB alone cannot efficiently resolve weakness (a) showing in Fig. \ref{fig:failurecases}.
\subsection{Dual-Scheme Guidance Module}
\begin{figure}[tbp]
\centering
\includegraphics[scale=0.3]{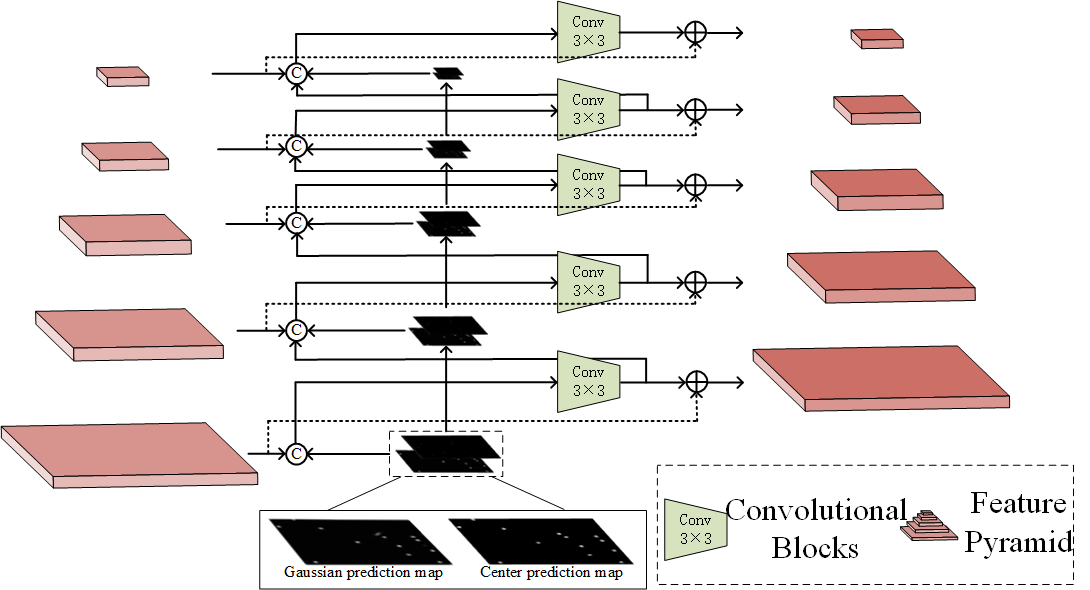}
\caption{Dual-Scheme Guidance Module}
\label{fig:dualscheme}
\end{figure}

Gaussian masks obtained in GGAB ($M^{G}\in\mathbb{R}^{H\times W}$) and the point masks obtained in the PRB ($M^{P}\in\mathbb{R}^{H\times W}$) contain information of each cell's center position. Therefore, the raw pyramid features can enhance cell location ability guided by these two masks. Through dual-scheme guidance, MSRNet can adapt to the cells of various appearances. Specifically, The prediction masks need to be processed by a series of max-pooling layers, then be concatenated into the feature map of the same size in pyramid features. At the same time, output features of the lower level layers will be processed by a downsampler and be concatenated to upper level input features. Convolutional blocks are then applied to the features after concatenation is completed, and the output feature maps have the same number of channels as the raw pyramid features. Finally, raw pyramid features need to be added to these feature maps, which are used for enable the fused feature map to retain the features in the original pyramid features. The framework of our dual-scheme guidance module is shown in Fig. \ref{fig:dualscheme}. Through this guidance, we obtain the input pyramid features of instance segmentation branch~\cite{DBLP:conf/nips/RenHGS15}.
\begin{equation}
\label{fusion_1}
F_{i}^{cat}=\left\{
\begin{aligned}
&Concat(F_{i}^{in}, M_{i}^{G}, M_{i}^{P}, F_{i-1}^{out}), &&i=1,2,3,4\\
&Concat(F_{i}^{in}, M^{G}, M^{P}), &&n=0
\end{aligned}
\right. 
\end{equation}
\begin{equation}
\label{fusion_ins}
\hfill F_{i}^{out}=f(F_{i}^{cat})+F_{i}^{in}, n=0,1,2,3,4
\end{equation}
where $F_{i}^{out}$ represents the feature map of the i$^{th}$ level of features in the output pyramid features, similarly, the $F_{i}^{in}$ represents the i$^{th}$ level of input pyramid features. The $M_{i}^{G}$ and $M_{i}^{P}$ represents $M^{G}$ and $M^{P}$ feature maps be processed by i downsamplers. 
\par Such processing allows the feature maps at each level to retain the detailed information while accept the useful information from dual-scheme prediction masks.
\subsection{Loss Functions}
By using a multi-task loss, we train our MSRNet in an end-to-end manner.
%%%%%
\begin{equation}
\label{g_pred}
\mathcal{L}_{G\_Pred}=\frac{1}{N\times H\times W}\sum_{n=1}^{N}\sum_{\substack{w=1\\h=1}}^{H\times W}\left|\left|\widehat{G}(w, h)-G(w, h)\right|\right|^{2}
\end{equation}
\par The gaussian mask prediction loss $\mathcal{L}_{G\_Pred}$ measures the distance between our predict gaussian masks and gaussian ground truths. $N$ represents the batch size. $G_{(w,h)}$ denotes the value of ground truths at the point $(w, h)$, while the $\widehat{G}_{(w,h)}$ denotes the value of predictions at the point $(w, h)$. 
\begin{equation}
\label{P_Loc}
\mathcal{L}_{P\_Loc}=-\frac{1}{N}\sum_{\substack{w=1\\h=1}}^{H\times W}(1-P_{(w,h)})^{\delta_{1}}(\widehat{P}_{(w,h)})^{\delta_{2}}log(1-\widehat{P}_{(w,h)})
\end{equation}
where $P_{(w,h)}$ indicates the value of ground truths at the point $(w, h)$, while the $\widehat{P}_{(w,h)}$ represents the value of predictions at the point $(w, h)$. $N$ represents the batch size, and $\delta_{1}$ and $\delta_{2}$ are hyper-parameters.
\begin{equation}
\label{HW_Reg}
\mathcal{L}_{HW\_Reg}=\frac{1}{n}\sum_{i=0}^{n}(\left|\widehat{c}_{n\_w}-\lfloor\frac{c_{n\_w}}{S}\rfloor\right|+\left|\widehat{c}_{n\_h}-\lfloor\frac{c_{n\_h}}{S}\rfloor\right|)
\end{equation}
where n represents the total number of cells in the input image. The $\widehat{c}_{n\_w}$ and $\widehat{c}_{n\_h}$ are width and length of each cell in our prediction. Only when the predicted cell center location is exactly the same as the actual cell location, the cell length and width will be calculated in the Loss function. The ${c_{n\_w}}$ and ${c_{n\_h}}$ represents the ground truths of height and width of cells. S represents the scale\_factor. Combining these two parts, the $\mathcal{L}_{P\_Reg}$ is as follows:
\begin{equation}
\label{P_Reg}
\mathcal{L}_{P\_Reg}=\beta_{1}\mathcal{L}_{P\_Loc} + \beta_{2}\mathcal{L}_{HW\_Reg}
\end{equation}
%%%%
\begin{table}[tbp]
  \caption{Comparison with previous SOTA instance segmentation methods. CE represents for the Cosine Embedding.}
  \centering
  \setlength{\tabcolsep}{0.3mm}{
  \begin{tabular}{c c c c c}
  \cmidrule(r){1-5}
  \multirow{2}{*}{Methods}
  &\multicolumn{2}{c}{DSB2018}   
  &\multicolumn{2}{c}{CA2.5} \\  

  \cmidrule(r){2-5}
                         &$AP_{50}$   &$AP_{75}$                             
                         &$AP_{50}$   &$AP_{75}$      \\

  \cmidrule(r){1-5}
  Mask R-CNN~\cite{He_2017_ICCV}  &69.9  &54.7 &87.5 &80.2\\
  \cmidrule(r){1-5}
  DCAN~\cite{Li_Liu_Lin_Xie_Ding_Huang_Tang_2020}  &51.9 &23.5 &72.4 &62.8 \\
  \cmidrule(r){1-5}
  CE~\cite{10.1007/978-3-030-00934-2_1}  &17.9 &3.4 &47.4 &24.4 \\
  \cmidrule(r){1-5}
  KG~\cite{DBLP:conf/miccai/YiWHQLHM19}  &71.6 &59.8 &- &- \\
  \cmidrule(r){1-5}
  SCNet~\cite{scnet_2021_aaai} &75.8 &61.2 &90.5 &83.2 \\
  \cmidrule(r){1-5}
  Ours &\textbf{77.0}  &\textbf{62.2} &\textbf{91.7}  &\textbf{84.0}\\
  \cmidrule(r){1-5}
  \end{tabular}}
 \label{result1}
\end{table}
\begin{table}[tbp]
  \caption{Comparison with previous SOTA instance segmentation methods on the SCIS dataset.The data in this table is from our code reproduction.}
  \centering
  \setlength{\tabcolsep}{2mm}{
  \begin{tabular}{c c c}
  \cmidrule(r){1-3}
  \multirow{2}{*}{Methods}
  &\multicolumn{2}{c}{SCIS} \\  
  \cmidrule(r){2-3}
                         &$AP_{50}$   &$AP_{75}$ \\

  \cmidrule(r){1-3}
  Mask R-CNN~\cite{He_2017_ICCV}  &18.2  &1.6\\
  \cmidrule(r){1-3}
  CM R-CNN~\cite{Cai_2019}  &26.5 &3.9\\
  \cmidrule(r){1-3}
  HTC~\cite{chen2019hybrid}  &28.6 &4.1\\
  \cmidrule(r){1-3}
  SCNet~\cite{scnet_2021_aaai}  &30.4 &4.2\\
  \cmidrule(r){1-3}
  Ours &\textbf{33.4}  &\textbf{4.9}\\
  \cmidrule(r){1-3}
  \end{tabular}}
 \label{result2}
\end{table}
%%%%%
$\mathcal{L}_{P\_Loc}$ represents the loss of cell centroid regression, while $\mathcal{L}_{HW\_Reg}$ is the loss used in height-width regression.
In summary, The multi-task loss is as follows:
\begin{equation}
\label{loss_overall}
\mathcal{L}=\lambda_{1}\mathcal{L}_{I}+\lambda_{2}\mathcal{L}_{G\_Pred}+\lambda_{3}\mathcal{L}_{P\_Reg}
\end{equation}
where $\mathcal{L}_{I}$ is similar to the Loss function used in the Cascade Mask R-CNN~\cite{Cai_2019}, $\mathcal{L}_{G\_Pred}$ is the loss used in predicting the gaussian masks in the GGAB, $\mathcal{L}_{P\_Reg}$ is the loss used in PRB. The $\lambda_{1}$, $\lambda_{2}$ and $\lambda_{3}$ are all hyper-parameters. 
\section{Experiments}
\subsection{Experimental Settings}
\subsubsection{Datasets}We evaluate MSRNet on three cell instance segmentation datasets including the 2018 Data Science Bowl (DSB2018), CA2.5~\cite{ca2.5_10.1007/978-3-030-87237-3_43} and Sartorius Cell Instance Segmentation (SCIS) datasets. DSB2018 dataset contains a total of 670 images, and the difficulty of this dataset mainly lies in the variety of image sizes, magnifications, imaging types and cell types. The CA2.5 dataset consists of 524 fluorescence images of 512×512 size, which contains some severely densely packed cell images with large differences in the brightness. The Sartorius Cell Instance Segmentation (SCIS) dataset is from a Kaggle‘s cell instance segmentation competition recently, which focus on neuronal cell instance segmentation. This dataset consists of a total of 606 images of 520×704 size. The images in this dataset are from LIVECell dataset, a large-scale dataset for label-free live cell segmentation~\cite{edlund2021livecell}.
\subsubsection{Implementation Details}
%%%%%
\begin{figure}[tbp]
\centering
\includegraphics[scale=0.21]{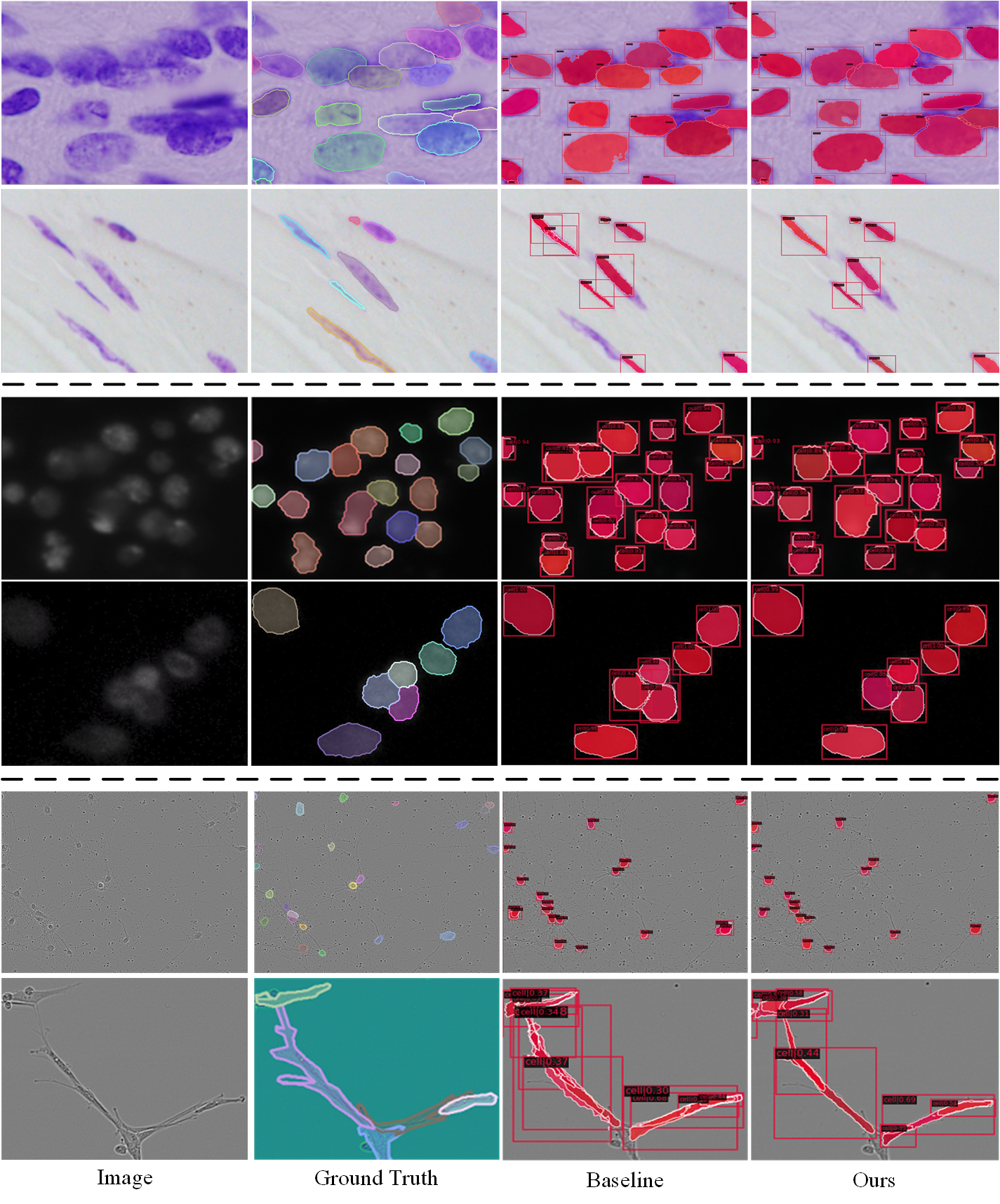}
\caption{The improvement of the instance segmentation result of our proposed method on all three datasets: the DSB2018, the CA2.5 and the SCIS datasets.}
\label{fig:comparetobase}
\end{figure}
We use ResNet101~\cite{He_2016_CVPR} as the backbone and experiment on three datasets. For the experiments on all datasets, images are resized to 512×512 and randomly divided into train, validation and test sets in the ratio of 6:2:2. Using Pytorch~\cite{DBLP:conf/nips/PaszkeGMLBCKLGA19} and MMDetection~\cite{mmdetection}, we conduct experiments on a 3090ti GPU with a batch\_size of 4 for 24 epochs using AdamW~\cite{DBLP:conf/iclr/LoshchilovH19} optimizer. The detection and segmentation results are evaluated with the standard COCO-style Average Precision (AP) metric.
\subsection{Results}
The performance of our method are compared with other SOTA instance segmentation methods, such as Mask R-CNN~\cite{He_2017_ICCV}, DCAN~\cite{Li_Liu_Lin_Xie_Ding_Huang_Tang_2020}, CosineEmbedding~\cite{10.1007/978-3-030-00934-2_1} and SCNet~\cite{scnet_2021_aaai}. Specially, due to the CA2.5 and SCIS dataset is very new, few result is available from other papers. So we use a series of SOTA methods on these datasets to get the baseline result. The results of the SOTA instance segmentation methods are shown in Tab. \ref{result1} and Tab. \ref{result2}. Obviously, our proposed method shows the best AP50 and AP75 scores compared with other SOTA methods. The visual comparisons of our proposed method with our baseline SCNet are shown in Fig. \ref{fig:comparetobase}, clearly, especially under dense conditions, our method performs better performance. The first line of this figure shows a simple case that our approach does not break the validity of the baseline method. Other rows contain some hard cases. For example, a single cell be recognized into two different cells (the cell in the upper left corner of the second row, and the cells on the left and right side of the fifth row, etc.). Besides, cells of the third and forth row show the densely packed cells be recognized into one cell. The last line shows an image that is very difficult for instance segmentation, for which our method has achieve some improvements compared to baseline. 
\par
\begin{figure}[tbp]
\centering
\includegraphics[scale=0.195]{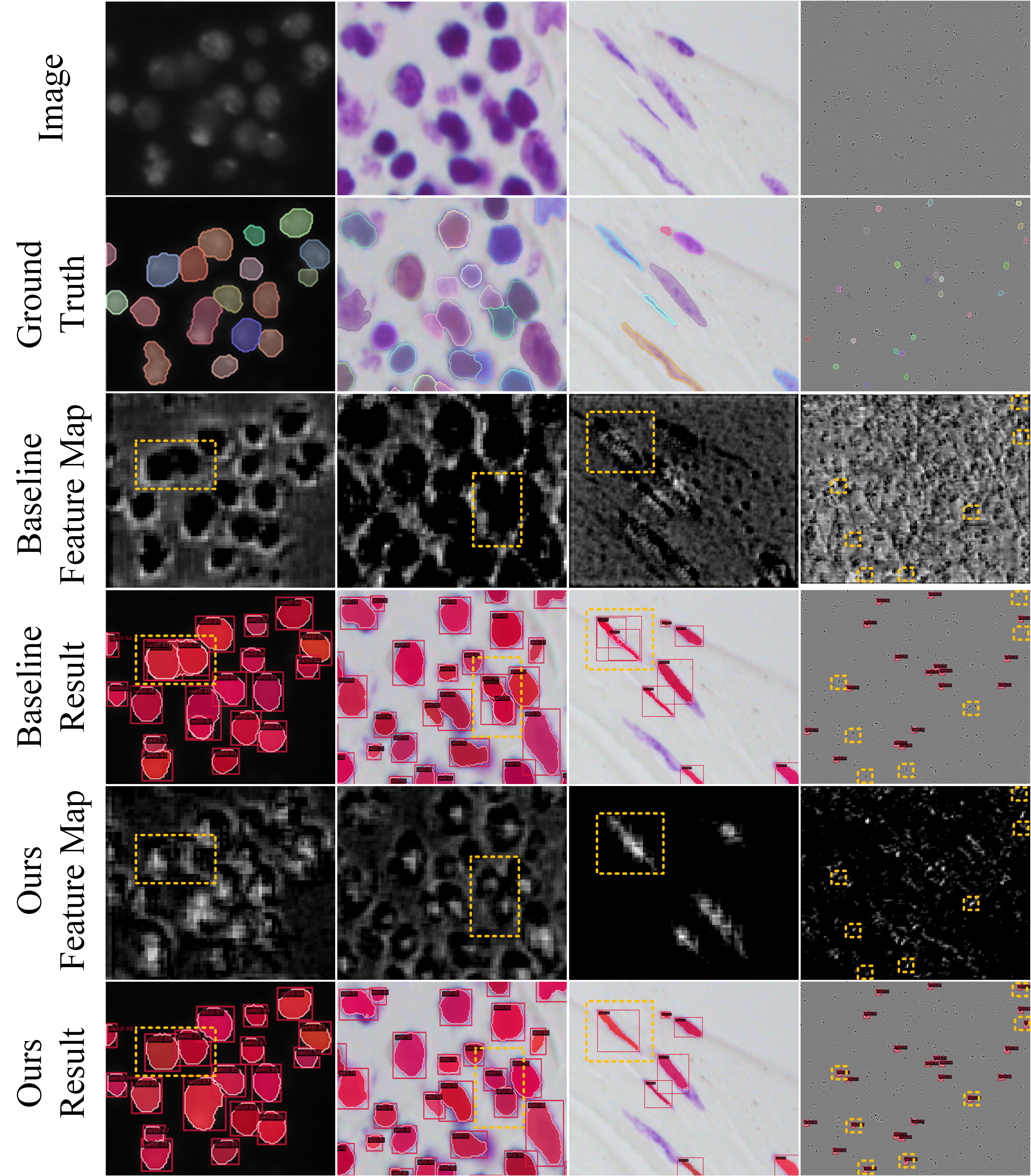}
\caption{Visualization results of feature maps}
\label{fig:comparetofeature}
\end{figure}

\begin{figure}[t]
\centering
\includegraphics[scale=0.17]{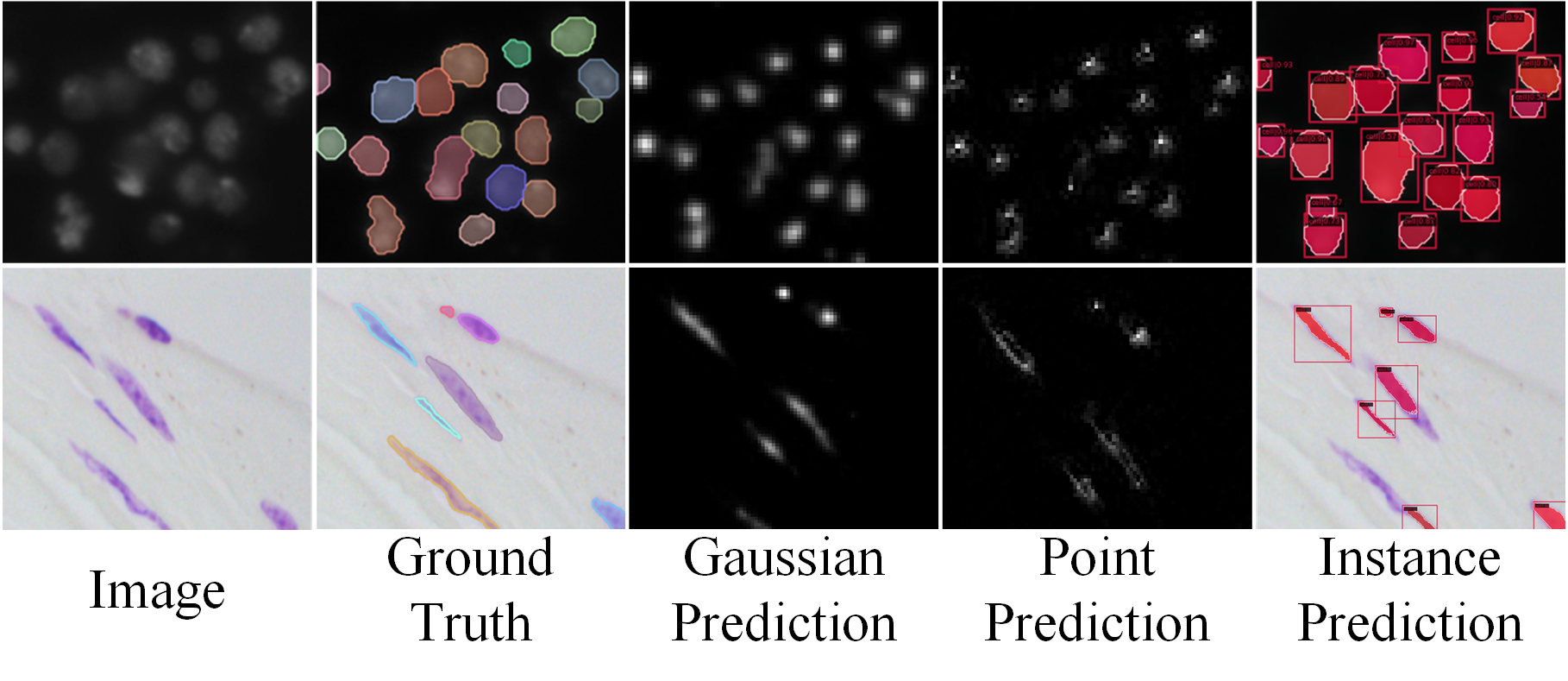}
\caption{Visualization of gaussian masks and point masks}
\label{fig:gaussianpoint}
\end{figure}

 Fig. \ref{fig:comparetofeature} shows the difference between MSRNet and baseline in terms of feature maps, from which we can directly explain the improvements of instance segmentation results. The first two columns show why our method can correctly recognize densely packed cells: in the original feature map, the feature regions of two cells are almost connected and indistinguishable, while our method's feature maps shows significant activation in each cell's central region, and the two activation do not overlap each other. MSRNet can use this feature to distinguish densely packed cells. The third column demonstrates why our method improves the recognition of elongated cells. MSRNet pays more attention to the central part of the cell, making it less likely be recognized into two cells. The fourth column demonstrates the improvement of the detection capability of our proposed MSRNet. MSRNet effectively reduces the probability of missing small cells by using multiple regression. In particular, fine-grained point-regression branch is useful for the detection and segmentation of small cells. Fig. \ref{fig:gaussianpoint} shows the gaussian prediction masks and point prediction masks. As can be seen from the figure, these two branches are able to correctly predict the location of each cell centers' position.
\begin{table}[tbp]
  \caption{Ablation study on the all three datasets. B represents the baseline, $m_{g}$ represents the GGAB, $m_{p}$ represents the PRB, $m_{d}$ is our dual-scheme guidance module. }
  \centering
  \setlength{\tabcolsep}{0.8mm}{
  \begin{tabular}{c c c c |c c c c c c}
  \cmidrule(r){1-10}
   \multirow{2}{*}{B} 
   &\multirow{2}{*}{+$m_{g}$} 
   &\multirow{2}{*}{+$m_{p}$} 
   &\multirow{2}{*}{+$m_{d}$}
  &\multicolumn{2}{c}{DSB2018}        
  &\multicolumn{2}{c}{CA2.5}   
  &\multicolumn{2}{c}{SCIS}   \\
  \cmidrule(r){5-10}
  & & &
                         &$AP_{50}$   &$AP_{75}$
                         &$AP_{50}$   &$AP_{75}$     
                         &$AP_{50}$   &$AP_{75}$      \\
  \cmidrule(r){1-10}
    \checkmark  &-  &-  &-  &75.8 &61.2 &90.5 &83.2 &30.4 &4.2 \\
  \cmidrule(r){1-10}
     \checkmark  &\checkmark &-  &- &76.6 &61.3 &91.2 &83.6 &31.5 &4.7 \\
  \cmidrule(r){1-10}
     \checkmark  &- &\checkmark  &- &76.2 &61.2 &90.8 &83.2 &31.2 &4.5 \\
  \cmidrule(r){1-10}
    \checkmark  &- &-  &\checkmark &76.0 &61.3 &91.0 &83.4 &31.5 &4.2\\
  \cmidrule(r){1-10}
     \checkmark  &\checkmark &\checkmark  &- &76.8 &61.5 &91.4 &83.8 &31.7 &4.9\\
  \cmidrule(r){1-10}
     \checkmark  &\checkmark &-  &\checkmark &77.0 &62.0 &91.6 &\textbf{84.2} &32.4 &4.8\\
  \cmidrule(r){1-10}
     \checkmark  &\checkmark &\checkmark  &\checkmark &\textbf{77. 0} &\textbf{62.2} &\textbf{91.7} &84. 0 &\textbf{33.4} &\textbf{4.9} \\
  \cmidrule(r){1-10}
  \end{tabular}}
 \label{ablation}
\end{table}
%%%%%
%%%%%
\begin{figure}[tbp]
\centering
\includegraphics[scale=0.16]{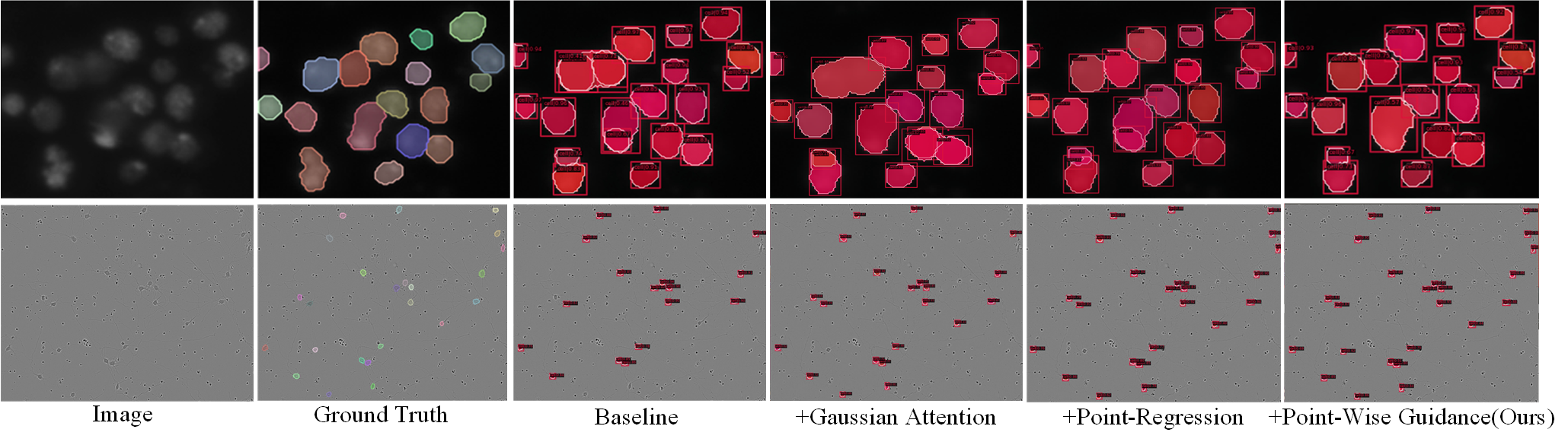}
\caption{Visualization results of each separated modules.}
\label{fig:ablation}
\end{figure}

%%%%%

\subsection{Ablation Study}
\subsubsection{Component-wise Analysis} A global component analysis was performed on our proposed MSRNet in order to show the effectiveness of it and the results are showed in Tab. \ref{ablation}. The baseline SCNet~\cite{scnet_2021_aaai} achieves the mask AP50 of 75.8\% on DSB2018 dataset. After GGAB is added, the mask AP50 improves 0.8\%. Then the dual-scheme guidance module is applied, the mask AP50 improves 0.1\%. After PRB is added, the mask AP50 improves 0.3\%. Overall, our proposed network obtains a total of 1.2\% AP50 improvement on the DSB2018 dataset, while obtains a total of 1.2\% improvement on the CA2.5\cite{ca2.5_10.1007/978-3-030-87237-3_43} dataset and a total of 3.0\% improvement on the SCIS dataset\cite{edlund2021livecell}.
\par From Fig. \ref{fig:ablation}, we can intuitively observe the effectiveness of each module. The GGAB mainly solves the weakness of single cell be recognized into multiple cells, while the PRB is primarily dedicated to solve the weakness of multiple cells be recognized into a single cell. The dual-scheme guidance module can guide the instance segmentation branch using the output of GGAB and PRB.
\subsubsection{Gaussian Guidance Attention Branch} The importance of GGAB has been clearly indicated in Tab. \ref{ablation}. After GGAB is applied on the DSB2018 dataset, the mask AP50 is improved from 75.8\% to 76.6\%. The mask AP50 on the other two datasets are also improved from 90.5\% to 91.2\% and 30.4\% to 31.5\%. To prove that our bi-directional hourglass module is effective, we compare the segmentation results before and after adding this module, as shown in Tab. \ref{w/o}. Tab. \ref{gaussian} shows the effect of changing the hyper-parameter gaussian radius. This table shows that changing the value of gaussian radius has a great impact on the performance of MSRNet. When the gaussian radius is too small, harsh conditions can lead to regression difficulties, while when the gaussian radius is too large, it will be more difficult for MSRNet to recognize densely packed cells.
%%%%%
\begin{table}[tbp]
  \caption{Ablation study of the gaussian guidance attention branch. H means the bi-directional hourglass module and C represents the dilation convolutional blocks. }
  \centering
  \setlength{\tabcolsep}{1mm}{
  \begin{tabular}{c c c c c c c}
  \cmidrule(r){1-7}
  \multirow{2}{*}{Methods}
  &\multicolumn{2}{c}{DSB2018}  
  &\multicolumn{2}{c}{CA2.5}  
  &\multicolumn{2}{c}{SCIS} \\  
  \cmidrule(r){2-7}
                         &$AP_{50}$   &$AP_{75}$  
                         &$AP_{50}$   &$AP_{75}$
                         &$AP_{50}$   &$AP_{75}$      \\

  \cmidrule(r){1-7}
  Ours w/o H  &76.8 &61.9 &91.5 &83.7 &33.1 &4.8 \\
  \cmidrule(r){1-7}
  Ours w/o C  &76.7 &61.7 &91.6 &83.8 &31.9 &4.5 \\
  \cmidrule(r){1-7}
  Ours  &\textbf{77.0} &\textbf{62.2} &\textbf{91.7} &\textbf{84.0} &\textbf{33.4} &\textbf{4.9}\\

  \cmidrule(r){1-7}
  \end{tabular}}
 \label{w/o}
\end{table}
%%%%%
%%%%%
\begin{table}[tbp]
  \caption{The difference of instance segmentation results made by using different gaussian radius on the 2018 Data Science Bowl dataset and CA2.5 dataset. }
  \centering
  \setlength{\tabcolsep}{2.6mm}{
  \begin{tabular}{c c c c c}
  \cmidrule(r){1-5}
  \multirow{2}{*}{Gaussian Radius} 
  &\multicolumn{2}{c}{DSB2018}   
  &\multicolumn{2}{c}{CA2.5} \\  
  \cmidrule(r){2-5}
                         &$AP_{50}$   &$AP_{75}$      
                         &$AP_{50}$   &$AP_{75}$      \\
  \cmidrule(r){1-5}
  1 / pixel  &75.9  &61.3 & 90.7 &83.3 \\
  \cmidrule(r){1-5}
  2 / pixel  &76.6 &61.5 &91.4 &83.7 \\
  \cmidrule(r){1-5}
  3 / pixel  &\textbf{77.0} &\textbf{62.0} &\textbf{91.6} &\textbf{84.2} \\
  \cmidrule(r){1-5}
  4 / pixel &76.8 &61.8 &91.5 &84.0 \\
  \cmidrule(r){1-5}
  5 / pixel &76.4 &61.0 &91.3 &83.8  \\
  \cmidrule(r){1-5}
  \end{tabular}}
 \label{gaussian}
\end{table}
%%%%%
\subsubsection{Dual-Scheme Guidance Branch}This branch can guide the instance segmentation branch by two types of masks from the output of GGAB and PRB. After applying this branch, the mask AP50 on DSB2018 dataset is improved from 76.6\% to 76.8\%. The mask AP50 on the other two datasets are also improved from 91.2\% to 91.6\% and 31.5\%to 32.4\%. 
\subsubsection{Point-Regression Branch} Tab. \ref{ablation} shows how much this branch can improve the performance of our proposed MSRNet. After this regression module is applied , the mask AP50 score on DSB2018 dataset has been improved from 76.8\% to 77.0\%. The mask AP50 scores on the other two datasets are also improved from 91.6\% to 91.7\% and 32.4\% to 33.4\%. Experimental results show that this fine-grained manner is effective, especially on the SCIS dataset, because the cell size in this dataset is smaller than those in other datasets.
\section{Conclusions}
In this paper, we propose a novel framework MSRNet for cell instance segmentation. By introducing the GGAB and the PRB, MSRNet can regress cell locations more accurately and be able to have a stronger perception of cell's central region features. In addition, the dual-scheme guidance module is introduced to guide the instance segmentation branch using the output of the other two branches. Our proposed MSRNet combines the idea of multi-scheme regression with cell instance segmentation, enabling the use of different views to understand the cellular characteristics and combining the information of each views for cell instance segmentation. Our MSRNet shows improvements over the baselines on the DSB2018, CA2.5 and SCIS datasets.

\begin{quote}
\bibliography{aaai23}

\begin{thebibliography}{36}
\providecommand{\natexlab}[1]{#1}

\bibitem[{Bolya et~al.(2019)Bolya, Zhou, Xiao et~al.}]{yolact-iccv2019}
Bolya, D.; Zhou, C.; Xiao, F.; et~al. 2019.
\newblock {YOLACT:} Real-Time Instance Segmentation.
\newblock In \emph{{IEEE/CVF} International Conference on Computer Vision},
  9156--9165.

\bibitem[{Bouyssoux, Fezzani, and
  Olivo{-}Marin(2022)}]{DBLP:conf/isbi/BouyssouxFO22}
Bouyssoux, A.; Fezzani, R.; and Olivo{-}Marin, J. 2022.
\newblock Cell Instance Segmentation Using Z-Stacks in Digital Cytology.
\newblock In \emph{{IEEE} International Symposium on Biomedical Imaging}, 1--4.

\bibitem[{Cai and Vasconcelos(2021)}]{Cai_2019}
Cai, Z.; and Vasconcelos, N. 2021.
\newblock Cascade {R-CNN:} High Quality Object Detection and Instance
  Segmentation.
\newblock \emph{{IEEE} Trans. Pattern Anal. Mach. Intell.}, 1483--1498.

\bibitem[{Chen et~al.(2019{\natexlab{a}})Chen, Pang, Wang
  et~al.}]{chen2019hybrid}
Chen, K.; Pang, J.; Wang, J.; et~al. 2019{\natexlab{a}}.
\newblock Hybrid Task Cascade for Instance Segmentation.
\newblock In \emph{{IEEE} Conference on Computer Vision and Pattern
  Recognition}, 4974--4983.

\bibitem[{Chen et~al.(2019{\natexlab{b}})Chen, Wang, Pang et~al.}]{mmdetection}
Chen, K.; Wang, J.; Pang, J.; et~al. 2019{\natexlab{b}}.
\newblock MMDetection: Open MMLab Detection Toolbox and Benchmark.
\newblock \emph{CoRR}.

\bibitem[{Cho et~al.(2014)Cho, van Merrienboer, G{\"{u}}l{\c{c}}ehre
  et~al.}]{DBLP:conf/emnlp/ChoMGBBSB14}
Cho, K.; van Merrienboer, B.; G{\"{u}}l{\c{c}}ehre, {\c{C}}.; et~al. 2014.
\newblock Learning Phrase Representations using {RNN} Encoder-Decoder for
  Statistical Machine Translation.
\newblock In Moschitti, A.; Pang, B.; and Daelemans, W., eds., \emph{{ACL}
  Proceedings of the 2014 Conference on Empirical Methods in Natural Language
  Processing}, 1724--1734.

\bibitem[{Edlund et~al.(2021)Edlund, Jackson, Khalid
  et~al.}]{edlund2021livecell}
Edlund, C.; Jackson, T.~R.; Khalid, N.; et~al. 2021.
\newblock LIVECell—A large-scale dataset for label-free live cell
  segmentation.
\newblock \emph{Nature methods}, 1038--1045.

\bibitem[{He et~al.(2017)He, Gkioxari, Doll{\'{a}}r et~al.}]{He_2017_ICCV}
He, K.; Gkioxari, G.; Doll{\'{a}}r, P.; et~al. 2017.
\newblock Mask {R-CNN}.
\newblock In \emph{{IEEE} International Conference on Computer Vision},
  2980--2988.

\bibitem[{He et~al.(2016)He, Zhang, Ren et~al.}]{He_2016_CVPR}
He, K.; Zhang, X.; Ren, S.; et~al. 2016.
\newblock Deep Residual Learning for Image Recognition.
\newblock In \emph{{IEEE} Conference on Computer Vision and Pattern
  Recognition}, 770--778.

\bibitem[{Huang et~al.(2021)Huang, Shen, Shen
  et~al.}]{ca2.5_10.1007/978-3-030-87237-3_43}
Huang, J.; Shen, Y.; Shen, D.; et~al. 2021.
\newblock CA\({}^{\mbox{2.5}}\)-Net Nuclei Segmentation Framework with a
  Microscopy Cell Benchmark Collection.
\newblock In de~Bruijne, M.; Cattin, P.~C.; Cotin, S.; et~al., eds.,
  \emph{{MICCAI} Medical Image Computing and Computer Assisted Intervention},
  Lecture Notes in Computer Science, 445--454.

\bibitem[{Li et~al.(2020{\natexlab{a}})Li, Wu, Peng et~al.}]{8733010}
Li, K.; Wu, Z.; Peng, K.; et~al. 2020{\natexlab{a}}.
\newblock Guided Attention Inference Network.
\newblock \emph{{IEEE} Trans. Pattern Anal. Mach. Intell.}, 2996--3010.

\bibitem[{Li et~al.(2020{\natexlab{b}})Li, Liu, Lin
  et~al.}]{Li_Liu_Lin_Xie_Ding_Huang_Tang_2020}
Li, S.; Liu, C.~H.; Lin, Q.; et~al. 2020{\natexlab{b}}.
\newblock Domain Conditioned Adaptation Network.
\newblock In \emph{The Thirty-Fourth {AAAI} Conference on Artificial
  Intelligence}, 11386--11393.

\bibitem[{Li, Zhang, and Chen(2018)}]{DBLP:conf/cvpr/LiZC18}
Li, Y.; Zhang, X.; and Chen, D. 2018.
\newblock CSRNet: Dilated Convolutional Neural Networks for Understanding the
  Highly Congested Scenes.
\newblock In \emph{{IEEE} Conference on Computer Vision and Pattern
  Recognition}, 1091--1100.

\bibitem[{Liu et~al.(2018)Liu, Qi, Qin et~al.}]{Liu_2018_CVPR}
Liu, S.; Qi, L.; Qin, H.; et~al. 2018.
\newblock Path Aggregation Network for Instance Segmentation.
\newblock In \emph{{IEEE} Conference on Computer Vision and Pattern
  Recognition}, 8759--8768.

\bibitem[{Loshchilov and Hutter(2019)}]{DBLP:conf/iclr/LoshchilovH19}
Loshchilov, I.; and Hutter, F. 2019.
\newblock Decoupled Weight Decay Regularization.
\newblock In \emph{7th International Conference on Learning Representations}.

\bibitem[{Ma et~al.(2019)Ma, Wei, Hong
  et~al.}]{DBLP:journals/corr/abs-1908-03684}
Ma, Z.; Wei, X.; Hong, X.; et~al. 2019.
\newblock Bayesian Loss for Crowd Count Estimation With Point Supervision.
\newblock In \emph{{IEEE/CVF} International Conference on Computer Vision},
  6141--6150.

\bibitem[{Oktay et~al.(2018)Oktay, Schlemper, Folgoc
  et~al.}]{DBLP:journals/corr/abs-1804-03999}
Oktay, O.; Schlemper, J.; Folgoc, L.~L.; et~al. 2018.
\newblock Attention U-Net: Learning Where to Look for the Pancreas.
\newblock \emph{CoRR}.

\bibitem[{Paszke et~al.(2019)Paszke, Gross, Massa
  et~al.}]{DBLP:conf/nips/PaszkeGMLBCKLGA19}
Paszke, A.; Gross, S.; Massa, F.; et~al. 2019.
\newblock PyTorch: An Imperative Style, High-Performance Deep Learning Library.
\newblock In Wallach, H.~M.; Larochelle, H.; Beygelzimer, A.; et~al., eds.,
  \emph{Advances in Neural Information Processing Systems 32: Annual Conference
  on Neural Information Processing Systems}, 8024--8035.

\bibitem[{Payer et~al.(2018)Payer, Stern, Neff
  et~al.}]{10.1007/978-3-030-00934-2_1}
Payer, C.; Stern, D.; Neff, T.; et~al. 2018.
\newblock Instance Segmentation and Tracking with Cosine Embeddings and
  Recurrent Hourglass Networks.
\newblock In Frangi, A.~F.; Schnabel, J.~A.; Davatzikos, C.; et~al., eds.,
  \emph{{MICCAI}Medical Image Computing and Computer Assisted Intervention},
  3--11.

\bibitem[{Ren et~al.(2015)Ren, He, Girshick et~al.}]{DBLP:conf/nips/RenHGS15}
Ren, S.; He, K.; Girshick, R.~B.; et~al. 2015.
\newblock Faster {R-CNN:} Towards Real-Time Object Detection with Region
  Proposal Networks.
\newblock In Cortes, C.; Lawrence, N.~D.; Lee, D.~D.; et~al., eds.,
  \emph{Advances in Neural Information Processing Systems 28: Annual Conference
  on Neural Information Processing Systems}, 91--99.

\bibitem[{Ren et~al.(2017)Ren, He, Girshick
  et~al.}]{DBLP:journals/corr/RenHG015}
Ren, S.; He, K.; Girshick, R.~B.; et~al. 2017.
\newblock Faster {R-CNN:} Towards Real-Time Object Detection with Region
  Proposal Networks.
\newblock \emph{{IEEE} Trans. Pattern Anal. Mach. Intell.}, 1137--1149.

\bibitem[{Schlemper et~al.(2018)Schlemper, Oktay, Chen
  et~al.}]{DBLP:journals/corr/abs-1804-05338}
Schlemper, J.; Oktay, O.; Chen, L.; et~al. 2018.
\newblock Attention-Gated Networks for Improving Ultrasound Scan Plane
  Detection.
\newblock \emph{CoRR}.

\bibitem[{Schmidt et~al.(2018)Schmidt, Weigert, Broaddus
  et~al.}]{10.1007/978-3-030-00934-2_30}
Schmidt, U.; Weigert, M.; Broaddus, C.; et~al. 2018.
\newblock Cell Detection with Star-Convex Polygons.
\newblock In Frangi, A.~F.; Schnabel, J.~A.; Davatzikos, C.; et~al., eds.,
  \emph{{MICCAI} Medical Image Computing and Computer Assisted Intervention},
  265--273.

\bibitem[{Sofiiuk, Barinova, and
  Konushin(2019)}]{DBLP:journals/corr/abs-1909-07829}
Sofiiuk, K.; Barinova, O.; and Konushin, A. 2019.
\newblock AdaptIS: Adaptive Instance Selection Network.
\newblock In \emph{{IEEE/CVF} International Conference on Computer Vision},
  7354--7362.

\bibitem[{Song et~al.(2018)Song, Huang, Ouyang
  et~al.}]{DBLP:conf/cvpr/Song0O018}
Song, C.; Huang, Y.; Ouyang, W.; et~al. 2018.
\newblock Mask-Guided Contrastive Attention Model for Person Re-Identification.
\newblock In \emph{{IEEE} Conference on Computer Vision and Pattern
  Recognition}, 1179--1188.

\bibitem[{Vu, Kang, and Yoo(2021)}]{scnet_2021_aaai}
Vu, T.; Kang, H.; and Yoo, C.~D. 2021.
\newblock SCNet: Training Inference Sample Consistency for Instance
  Segmentation.
\newblock In \emph{{AAAI} Conference on Artificial Intelligence, {AAAI} 2021,
  Thirty-Third Conference on Innovative Applications of Artificial
  Intelligence, {IAAI} 2021, The Eleventh Symposium on Educational Advances in
  Artificial Intelligence, {EAAI} 2021, Virtual Event, February 2-9, 2021},
  2701--2709.

\bibitem[{Wang et~al.(2020{\natexlab{a}})Wang, Kong, Shen
  et~al.}]{wang2020solo}
Wang, X.; Kong, T.; Shen, C.; et~al. 2020{\natexlab{a}}.
\newblock {SOLO:} Segmenting Objects by Locations.
\newblock In Vedaldi, A.; Bischof, H.; Brox, T.; et~al., eds., \emph{Computer
  Vision - {ECCV} 16th European Conference}, Lecture Notes in Computer Science,
  649--665.

\bibitem[{Wang et~al.(2020{\natexlab{b}})Wang, Zhang, Kong
  et~al.}]{wang2020solov2}
Wang, X.; Zhang, R.; Kong, T.; et~al. 2020{\natexlab{b}}.
\newblock SOLOv2: Dynamic and Fast Instance Segmentation.
\newblock In Larochelle, H.; Ranzato, M.; Hadsell, R.; et~al., eds.,
  \emph{Advances in Neural Information Processing Systems 33: Annual Conference
  on Neural Information Processing Systems.}

\bibitem[{Xie et~al.(2020)Xie, Sun, Song et~al.}]{Xie_2020_CVPR}
Xie, E.; Sun, P.; Song, X.; et~al. 2020.
\newblock PolarMask: Single Shot Instance Segmentation With Polar
  Representation.
\newblock In \emph{{IEEE/CVF} Conference on Computer Vision and Pattern
  Recognition}, 12190--12199.

\bibitem[{Xu et~al.(2015)Xu, Ba, Kiros et~al.}]{DBLP:conf/icml/XuBKCCSZB15}
Xu, K.; Ba, J.; Kiros, R.; et~al. 2015.
\newblock {JMLR}Show, Attend and Tell: Neural Image Caption Generation with
  Visual Attention.
\newblock In Bach, F.~R.; and Blei, D.~M., eds., \emph{Proceedings of the 32nd
  International Conference on Machine Learning}, 2048--2057.

\bibitem[{Yang et~al.(2016)Yang, Yang, Dyer
  et~al.}]{DBLP:conf/naacl/YangYDHSH16}
Yang, Z.; Yang, D.; Dyer, C.; et~al. 2016.
\newblock Hierarchical Attention Networks for Document Classification.
\newblock In Knight, K.; Nenkova, A.; and Rambow, O., eds., \emph{{NAACL} The
  2016 Conference of the North American Chapter of the Association for
  Computational Linguistics: Human Language Technologies}, 1480--1489.

\bibitem[{Yi et~al.(2018{\natexlab{a}})Yi, Wu, Hoeppner
  et~al.}]{DBLP:conf/isbi/YiWHM18}
Yi, J.; Wu, P.; Hoeppner, D.~J.; et~al. 2018{\natexlab{a}}.
\newblock Pixel-wise neural cell instance segmentation.
\newblock In \emph{{IEEE} International Symposium on Biomedical Imaging},
  373--377.

\bibitem[{Yi et~al.(2019{\natexlab{a}})Yi, Wu, Huang
  et~al.}]{DBLP:conf/isbi/YiWHQHM19}
Yi, J.; Wu, P.; Huang, Q.; et~al. 2019{\natexlab{a}}.
\newblock Context-Refined Neural Cell Instance Segmentation.
\newblock In \emph{{IEEE} International Symposium on Biomedical Imaging},
  1028--1032.

\bibitem[{Yi et~al.(2019{\natexlab{b}})Yi, Wu, Huang
  et~al.}]{DBLP:conf/miccai/YiWHQLHM19}
Yi, J.; Wu, P.; Huang, Q.; et~al. 2019{\natexlab{b}}.
\newblock Multi-scale Cell Instance Segmentation with Keypoint Graph Based
  Bounding Boxes.
\newblock In Shen, D.; Liu, T.; Peters, T.~M.; et~al., eds., \emph{{MICCAI}
  Medical Image Computing and Computer Assisted Intervention}, 369--377.

\bibitem[{Yi et~al.(2018{\natexlab{b}})Yi, Wu, Jiang
  et~al.}]{DBLP:conf/eccv/YiWJHM18}
Yi, J.; Wu, P.; Jiang, M.; et~al. 2018{\natexlab{b}}.
\newblock Instance Segmentation of Neural Cells.
\newblock In Leal{-}Taix{\'{e}}, L.; and Roth, S., eds., \emph{Computer Vision
  - {ECCV} 2018 Workshops}, Lecture Notes in Computer Science, 395--402.

\bibitem[{Zhang et~al.(2016)Zhang, Zhou, Chen
  et~al.}]{DBLP:conf/cvpr/ZhangZCGM16}
Zhang, Y.; Zhou, D.; Chen, S.; et~al. 2016.
\newblock Single-Image Crowd Counting via Multi-Column Convolutional Neural
  Network.
\newblock In \emph{{IEEE} Conference on Computer Vision and Pattern
  Recognition}, 589--597.

\end{thebibliography}
\end{quote}

\end{document}